\documentclass{article} % For LaTeX2e
\usepackage{iclr2020_conference,times}

\usepackage{amsmath,amsfonts,bm}

\def\eqref#1{equation~\ref{#1}}
\def\1{\bm{1}}

\DeclareMathAlphabet{\mathsfit}{\encodingdefault}{\sfdefault}{m}{sl}
\SetMathAlphabet{\mathsfit}{bold}{\encodingdefault}{\sfdefault}{bx}{n}

\usepackage{hyperref}
\usepackage{url}
\usepackage{graphicx}
\usepackage{algorithm, algorithmic}
\usepackage{mathrsfs}
\usepackage{ulem} 

\title{Unsupervised Few-shot Learning via Self-supervised Training}

\author{Zilong Ji \\
State Key Laboratory of Cognitive \\
Neuroscience \& Learning,\\
Beijing Normal University, Beijing, China \\
\texttt{jizilong@mail.bnu.edu.cn} \\
\And
Xiaolong Zou, Tiejun Huang\\
School of Electronics Engineering \\ 
and Computer Science, \\
Peking University, Beijing, China\\
\texttt{xiaolz, tjhuang@pku.edu.cn} 
\And
Si Wu \\
School of Electronics Engineering \\
and Computer Science, \\
IDG/McGovern Institute for Brain Research, \\
Peking University, Beijing, China\\
\texttt{siwu@pku.edu.cn} 
}

\iclrfinalcopy % Uncomment for camera-ready version, but NOT for submission.
\begin{document}

\maketitle

\begin{abstract}
Learning from limited exemplars (few-shot learning) is a fundamental, unsolved problem that has been laboriously explored in the machine learning community. However, current few-shot learners are mostly supervised and rely heavily on a large amount of labeled examples. Unsupervised learning is a more natural procedure for cognitive mammals and has produced promising results in many machine learning tasks. In the current study, we develop a method to learn an unsupervised few-shot learner via self-supervised training (UFLST), which can effectively generalize to novel but related classes.
The proposed model consists of two alternate processes, progressive clustering and episodic training. The former generates pseudo-labeled training examples for
constructing episodic tasks; and the later trains the few-shot learner using the generated episodic tasks which further optimizes the feature representations of data. The two processes facilitate with each other, and eventually produce a high quality few-shot learner. 
Using the benchmark dataset Omniglot and Mini-ImageNet, we show that our model outperforms other unsupervised few-shot learning methods.
Using the benchmark dataset Market1501, we further demonstrate the feasibility of our model to a real-world application on person re-identification.
\end{abstract}

\section{Introduction}
Few-shot learning, which aims to accomplish a learning task by using very few training examples, is receiving increasing attention in the machine learning community. The challenge of few-shot learning lies on that traditional techniques such as fine-tuning would normally incur overfitting~\citep{wang2018low}. To overcome this difficulty, a \textit{set-to-set} meta-learning(episodic learning) paradigm was proposed~\citep{vinyals2016matching}. In such a paradigm, 
the conventional mini-batch training is replaced by the episodic training, in term of that a batch of episodic tasks, 
each of which having the same setting as the testing environment,
are presented to the learning model; and in each episodic task,  the model learns to predict the classes of unlabeled points (the query set) using very few labeled examples (the support set). By this, the learning model acquires the transferable knowledge (optimized feature representations) across tasks, and due to the consistency between the training and testing environments, the model is able to generalize to novel but related tasks.
Although this set-to-set few-shot learning paradigm has made great progress, in its current supervised form, it requires a large number of labeled examples for constructing episodic tasks, which is often infeasible or too expensive in practice. So, can we build up a few-shot learner in the paradigm of episodic training using only unlabeled data? 

It is well-known that humans have the remarkable ability to learn 
a concept when given only several exposures to its instances, for example, young children can effortlessly learn and generalize the concept of ``giraffe" after seeing a few pictures of giraffes. While the specifics of the human learning process are complex (trial-based, perpetual, multi-sourced, and simultaneous for multiple tasks) and yet to be solved, previous works agree that its nature is progressive and unsupervised in many cases~\citep{dupoux2018cognitive}. 
Given a set of unlabeled items, humans are able to organize them into different clusters by comparing one with another. The comparing or associating process follows a \textit{coarse-to-fine} manner. At the beginning of learning, humans tend to group items based on fuzzy-rough knowledge such as color, shape or size. Subsequently, humans build up associations between items using more fine-grained knowledge, i.e., stripes of images, functions of items or other domain knowledge. Furthermore, humans can extract representative representations across categories and apply this capability to learn new concepts~\citep{wang2014vernier}.

\begin{figure}[t] 
\begin{center}
\includegraphics[width=1.0\linewidth]{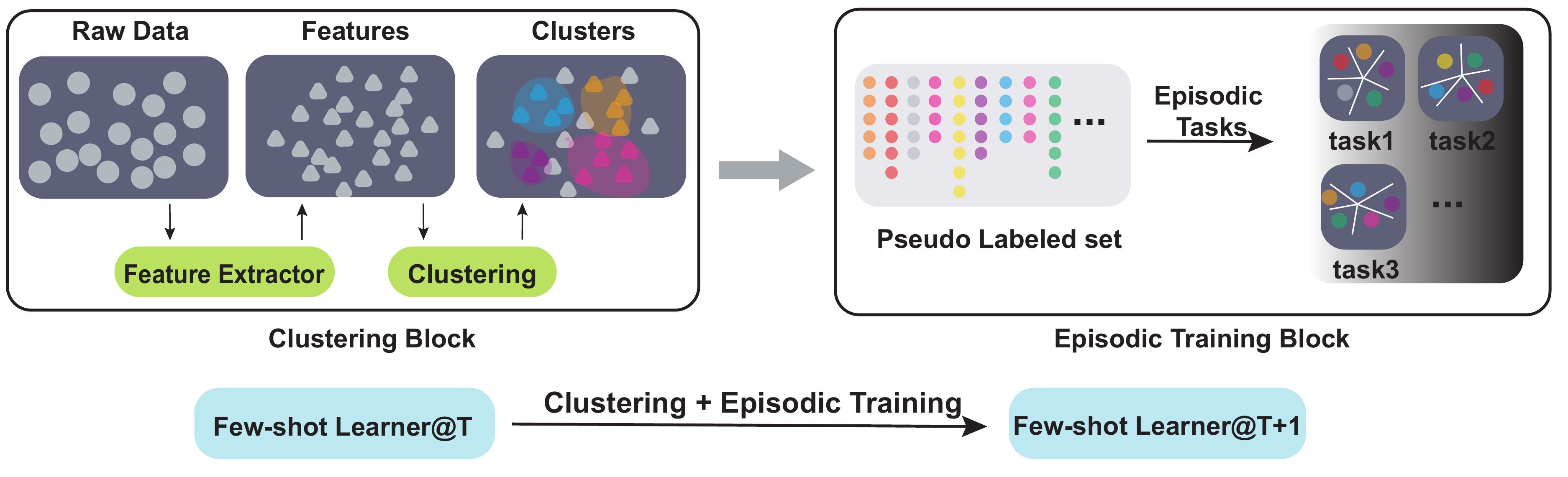}
\end{center}
  \caption{The scheme of our model UFLST, which consists of two alternate processes: clustering and episodic training. At each round, unlabeled data points are clustered based on extracted features, and pseudo labels are assigned according to cluster identities. After clustering, a set of
  episodic tasks are constructed by sampling from the pseudo-labeled data, and the few-shot learner is trained, which further optimizes feature representations. The two processes are repeated.}
  \label{fig-model}
\end{figure}

In the present study, inspired by the unsupervised and progressive characteristics of human learning, we propose an unsupervised model for few-shot learning via a self-supervised training procedure (UFLST).
Different from previous unsupervised learning methods, our model integrates unsupervised learning and episodic training into a unified framework, which
facilitates feature extraction and model training iteratively.
Basically, we adopt the episodic training paradigm, taking advantage of its capability of extracting transferable knowledge across tasks, but we use an unsupervised strategy to construct episodic tasks. Specifically, we apply progressive clustering to  
generate pseudo labels for unlabeled data, and 
this is done alternatively with feature optimization via few-shot learning in an iterative manner (Fig.~\ref{fig-model}). Initially, unlabeled data points are assigned into several clusters, and we sample a few training examples from each cluster together with their pseudo labels (the identities of clusters) to construct a set of episodic tasks having the same setting as the testing environment. We then train the few-shot learner using the constructed episodic tasks and obtain improved feature representations for the data. In the next round, we use the improved features to re-cluster data points, generating new pseudo labels and constructing new episodic tasks, and train the few-shot learner again. The above two steps are repeated till a stopping criterion is reached. After training, we expect that the few-shot learner has acquired the transferable knowledge (the optimized feature representations) suitable for a novel task of the same setting as in the episodic training. Using benchmark datasets, we demonstrate that our model outperforms other unsupervised few-shot learning methods and approaches to the performances of fully supervised models. 

\section{Related Works}\label{related-works}

In the paradigm of episodic training, few-shot learning algorithms can be divided into two main categories: ``learning to optimize" and ``learning to compare". The former aims to develop a learning algorithm which can adapt to a new task efficiently using only few labeled examples or with few steps of parameter updating~\citep{finn2017model,ravi2016optimization,mishra2017simple, rusu2018meta, nichol2018reptile, andrychowicz2016learning}, and 
the latter aims to learn a proper embedding function, so that prediction is based on the distance (metric) of a novel example to the labeled instances~\citep{vinyals2016matching, snell2017prototypical, ren2018meta, sung2018learning, liu2018learning}. In the present study, we focus on the ``learning to compare" framework, although the other framework can also be integrated into our model.

Only very recently, people have tried to develop unsupervised few-shot learning models. \citet{hsu2018unsupervised} proposed a method called CACTUs, which uses progressive clustering to leverage feature representations before carrying out episodic training. This is different from our model, in term of that we carry out progressive clustering and episodic training concurrently, 
and the two processes facilitate with each other.
\citet{khodadadeh2018unsupervised} proposed a method called UMTRA, which utilizes the statistical diversity properties and domain-specific augmentations to generate training and validation data. \citet{antoniou2019assume} proposed a similar model called AAL, which uses data augmentations of the unlabeled support set to generate the query data. Both methods are different from our model, in term of that we use a progressive clustering strategy to generate pseudo labels for constructing episodic tasks.    

The idea of self-supervised training is to artificially generate 
pseudo labels for unlabeled data, which is useful when supervisory signals are not available or too expensive~\citep{de1994learning}.
Progressive (deep) clustering is a promising method for self-supervised training, which aims to optimize feature representations and pseudo labels (cluster assignments) in an iterative manner.
This idea was first applied in NLP tasks, which tries to self-train a two-phase parser-reranker system using unlabeled data~\citep{mcclosky2006effective}. 
\citet{xie2016unsupervised} proposed a Deep Embedded Clustering network to jointly learn cluster centers and network parameters. 
\citet{caron2018deep} further proposed strategies to solve the 
degenerated solution problem in progressive clustering.
\citet{fan2018unsupervised} and \citet{song2018unsupervised} applied the progressive clustering idea to the person re-identification task, both of which aim to transfer the extracted feature representations to an unseen domain. 
None of these studies have integrated progressive clustering 
and episodic training in few-shot learning as we do in this work.

\section{Method}
In this section, we describe the model UFLST in detail.
Let us first introduce some notations. 
Denote the model at the training round $t$ as $M^t$, $\{x_i\}$ the unlabeled dataset with the number
of examples $N$, and $\{z_i\}^t$ is the corresponded feature vector with dimentionality $D$, which is given by $f_{\theta^{t}}: x \xrightarrow{M^t} z$, where $f_{\theta^{t}}$ representing the feature extracter and $\theta^t$ the training parameters of $M^t$. 
$\{\widetilde{z}_i\}^t$ and $\{\widetilde{x}_i\}^t$ represent,
respectively, the selected features and unlabeled data after removing outliers from the clustering results, and $\{\widetilde{y}_i\}^t$ the corresponding pseudo labels. 

\subsection{Progressive Clustering}

\subsubsection{K-reciprocal Jaccard Distance}
\begin{figure}[htbp]
\begin{center}
\includegraphics[width=1.0\linewidth]{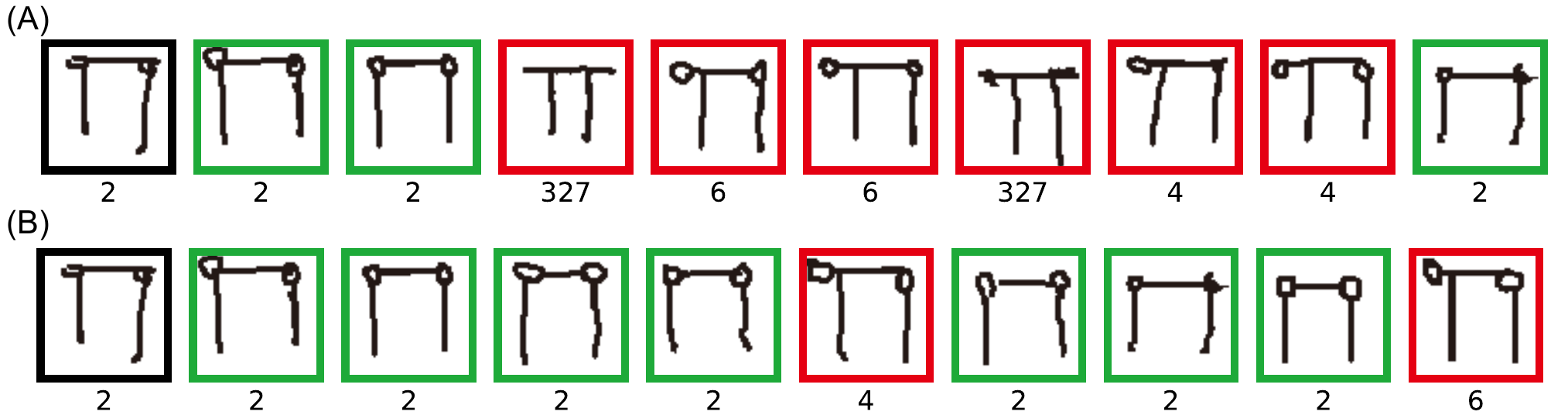}
\end{center}
  \caption{Comparing the performances of KRJD and the Euclidean metric. Top 10 neighbours of a chosen query character from Omniglot are shown. Black box: the query character. Green box: the positive characters in the neighbourhood of the query character. Red box: the negative characters in the neighbourhood of query character. (A) The ranking result using Euclidean metric. (B) The ranking result using KRJD. The number under each image represents its true class. KRJD outperforms the Euclidean metric, in term of it includes more positive examples in the ranking list.}
  \label{fig-rerankcomp}
\end{figure}

To cluster unlabeled data points,
we adopt the k-reciprocal Jaccard distance (KRJD) metric to measure
the distance between data points~\citep{qin2011hello, zhong2017re},
and it is done in the feature space rather than the raw pixel space.
First, we calculate the k-reciprocal nearest neighbours of each feature point, which are given by,
\begin{equation}\label{equ-krnn}
    R(z, k) = \{z_j\ | (z_j \in N(z,k)) \cap (z \in N(z_j,k))\},
\end{equation}
where $N(z,k)$ denotes the $k$ nearest neighbours of $z$. $R(z, k)$ imposes that $z$ and each element of $R(z,k)$ are mutually
$k$ nearest neighbours of each other. Second, we compute KRJD between two feature points, which is given by
\begin{equation}\label{equ-krjd}
    J_{ij} = 1-\frac{|R(z_i,k) \cap R(z_j,k)|}{|R(z_i,k) \cup R(z_j,k)|}.
\end{equation}
Compared to the Euclidean distance, KRJD takes into account the reciprocal relationship between data points, and hence is a stricter rule measuring whether two feature points matches or not. We find that KRJD is crucial to our model, which outperforms 
the Euclidean metric as
demonstrated in Fig.~\ref{fig-rerankcomp}.

\subsubsection{Density-based spatial clustering algorithm}
For the clustering strategy, we choose the density-based spatial clustering algorithm (DBSCAN)~\citep{ester1996density}, which performs
better than other methods in our model. 
The reasons are:
1) other clustering methods such as k-means and hierarchical clustering are useful to find spherical shaped or convex clusters in the embedding space, while DBSCAN works well for arbitrarily shaped clusters; 2) DBSCAN can detect outliers as noise points, which is very useful at the beginning of training when the distribution of data points in the embedding space is highly noisy; 3) DBSCAN does not need to specify the number of clusters to be generated, which is appealing for unsupervised learning. 
After removing noisy points (outliers) as done in DBSCAN, pseudo labels (i.e., cluster identities) of data points $\{\widetilde{x}_i\}^t$ can be expressed as,
\begin{equation}
    \{\widetilde{y}_i\}^t=DBSCAN(\epsilon,ms,J),
\end{equation}
where $\epsilon$ denotes the maximum distance for two points to be considered as in the same neighborhood, $ms$ the minimum number of points huddled together for a region to be considered as dense, and $J$ the KRJD matrix. The value of $\epsilon$ relies on the cluster density of feature points, which is set to be the mean distance of top $P$ minimum distances in KRJD in this study.

\subsection{Episodic training}

\subsubsection{Constructing episodic tasks}\label{episodic-tasks}
After each round of clustering, we construct a number of episodic tasks, denoted as $\mathcal{T}=\{T_1, T_2, ..., T_S\}$ with $S$ the number of tasks, from the pseudo-labeled data set
$\{\widetilde{x}_i, \widetilde{y}_i\}^t$. 
For each episodic task, we randomly sample
$N_C$ classes and $N_E$ examples per class.
Notably, the setting of each episodic task follows that of the test environment to be performed after training.

We will apply two different ways to implement few-shot learning (see below). One uses the prototype loss, which aims to learn the prototypes of each class and discriminate a novel example based on its distances to the prototypes. In this case, we further split $N_E$ into a support set $S$ and a query set $Q$, i.e., $N_E=N_S+N_Q$. Following~\citet{snell2017prototypical}, 
we choose a larger value of $N_C$ in training than that in testing,
but keeps $N_S$ the same.
The other way to implement few-shot learning is to use
the triplet loss or the hardtriplet loss, which separates examples from different classes with a positive margin $m$. 
In this case, no splitting support and query data is needed.
To mine hard negative examples in triplets, we also use a larger $N_C$ in training than that in testing. 

\subsubsection{Loss Functions}\label{losscompare}
Two types of loss functions are used in the present study, and
both of them are in the framework of  ``learning to compare" and contribute to simple inductive bias of our model.
One is the prototype loss,
which is written as
\begin{equation}\label{equ-3}
    L_{proto}(z, {c_p};\theta) = \frac{\exp(-\Vert{z-c_p}\Vert_2^2)}{\sum_{k}\exp(-\Vert{z-c_{k}}\Vert_2^2)},
\end{equation}
where $c_k$ is the prototype of class $k$ given by
$ c_k = \sum_{z_i \in S_k}(z_i)/S_k$, and
$z$ a query point. In implementation, we choose to minimize the negative log value of Eq.~\ref{equ-3}, i.e.,
$L_{proto}^{\log}(z, {c_k};\theta)=-\log L_{proto}(z, {c_k};\theta)$, 
as the log value better reflects the geometry of
the loss function, making it easy to select a learning rate to minimize the loss function.

The other is the triplet loss~\citep{weinberger2009distance}, which has been widely used in face recognition and image retrieval. The triplet loss $L_{triplet}$ consists of several triplets, each of which includes a query feature $z$, a positive feature $z^+$ and a negative feature $z^-$, and is written as 
\begin{equation}\label{equ-0}
    L_{triplet}(z,z^+,z^-;\theta)=\max(0, \Vert{z-z^+}\Vert_2^2-\Vert{z-z^-}\Vert_2^2+m),
\end{equation}
where $m$ controls the margin of two classes, and the hinge term plays the role of correcting triplets, so that the difference between the similarities of positive and negative examples to the query point is larger than a margin $m$.  
However, in the above form, positive pairs in those ``already correct" triplets will no longer be pulled together due to the hard cutoff. We therefore replace the hinge term by a soft-margin formulation, which gives 
\begin{equation}\label{equ-1}
    L_{triplet-SM}(z,z^+,z^-;\theta)=\log\left[1+\exp(\Vert{z-z^+}\Vert_2^2-\Vert{z-z^-}\Vert_2^2+m)\right].
\end{equation}
Eq.~\ref{equ-1} is similar to Eq.~\ref{equ-0}, but it decays exponentially instead of having a hard cutoff and tends to be numerically more stable~\citep{hermans2017defense}.

We find that in general our model achieves a better performance using the prototype loss than using the triplet loss. However, by including hard example mining when constructing triplets, referred to as the hardtriplet loss hereafter, the model performance is improved significantly and becomes better that  using the prototype loss.  
The pseudo code of our model is summarized in Algorithm \ref{algo}.

\begin{algorithm}[t]
	\renewcommand{\algorithmicrequire}{\textbf{Input:}} 
	\renewcommand{\algorithmicensure}{\textbf{Output:}} 
	\caption{Unsupervised Few-shot Learning via Self-supervised Training (UFLST)}
	\label{algo}
	\begin{algorithmic}[1]
	  \REQUIRE Unlabeled dataset $\{x_i\}$, initialized model parameters $\theta^0$
	  \ENSURE Trained model parameters $\theta^{T}$
	  \STATE $t=0$
	  \REPEAT
	  \STATE \textbf{Clustering:}
	  \STATE Extracting features $\{z_i\}^t$ of $\{x_i\}$ using the feature extractor $f_{\theta^t}$.
	  \STATE Computing K-reciprocal nearest neighbours $R(z_i, k)^t$ of each $z_i$.
	  \STATE Calculating Jaccard distance matrix $J$ based on $\{R(z_i, k)\}^t$.
	  \STATE Clustering data using DBSCAN
	  and generating pseudo labels $\{y_i\}^t$.
	  \STATE Removing outliers and obtaining the 
	  pseudo-labeled dataset $\{\widetilde{x}_i, \widetilde{y}_i\}^t$.
	  \STATE \textbf{Episodic Training:}
	   \STATE $s=0$
	  \REPEAT 
	  \STATE Constructing a episodic task $\mathcal{T}^s$ by randomly sampling $N_C$ classes with $N_E$ examples per class from $\{\widetilde{x}_i, \widetilde{y}_i\}^t$.
	  \STATE Updating model parameters $\theta^t$ by training the few-shot learner on $\mathcal{T}^s$.
	  \STATE $s=s+1$
	  \UNTIL{$s=S$}
	  \STATE $t=t+1$
	  \UNTIL{$t=T$}
	\end{algorithmic} 
	\label{algorithm1}
\end{algorithm}

\section{Experiments}

\subsection{Datasets}

We evaluate our model on three benchmark datasets, which are Omniglot~\citep{lake2015human}, Mini-ImageNet~\citep{vinyals2016matching} and Market1501~\citep{zheng2015scalable}.

\textbf{Omniglot}  contains 1623 different handwritten characters from 50 different alphabets. There are 20 examples in each class and each of them was drawn by a different human subject via Amazon's Mechanical Turk. We split data into two parts: 1200 characters for training and 423 for testing, but we did not augment data with rotations (this is unnecessary in our model), and instead of resizing images to $28 \times 28$, we resized them to $32 \times 32$.   

\textbf{Mini-Imagenet} is derived from the ILSVRC-12 dataset. We follow the data splits proposed by Ravi and Larochelle~\cite{ravi2016optimization}, which has a different set of 100 classes including 64 for training, 16 for validating, and 20 for testing compared to the spilt by ~\citet{vinyals2016matching}. Each class contains 600 colored images of size $84 \times 84$.

\textbf{Market1501} is a person re-identification (Re-ID) dataset containing 32668 images with 1501 identities captured from 6 cameras. The dataset is split into three parts: 12936 images with 751 identities forming the training set, 19732 images with 750 identities forming the gallery set, and another 3368 images forming the query set. All images were resized to $256 \times 128$. Except normalization, no other pre-processing was applied.

\subsection{Implementation Details}\label{imple-detail}

When training on the Omniglot and the Mini-ImageNet dataset, we chose the model architecture to be the same as that in~\citet{vinyals2016matching} , which consists of four stacked layers. Each layer comprises 64-filter $3\times3$ convolution, followed by a batch normalization, a ReLU nonlinearity, and $2\times2$ max-pooling. 
When training on the market1501 dataset, due to 
high variances of pose and luminance, we chose to use a highly expressive model~\citep{xiong2018towards}, which consists of a Resnet50 pretrained on ImageNet as a backbone, and a batch normalization layer after the global max-pooling layer to prevent overfitting. Our evaluation protocol on market1501 is different from that in \citet{zheng2015scalable}, where they reported Cumulative Matching Characteristic (CMC) and the mean average precision (mAPs); while we consider the performance of 1-shot learning, which mimics the typical single query condition in a person Re-ID task. 

When training with the triplet loss, we set
the margin between positive and negative examples to be $0.5$,
and the number of training rounds $T=20$. To avoid overfitting, the model is fine-tuned for only $50$ epochs in each round. We used Adam with momentum to update the model parameters, and the learning rate is set to $0.005$ with an exponential decay after $25$ epochs. The mini-batch size is $128$, which consists of $32$ classes and $4$ examples per class in each episodic task. When constructing triplets with hard example mining, we didn't mine hard negative examples across the whole dataset which is infeasible, rather we did only in the current episodic task.
When training with the prototype loss, we used more classes (higher “way”) during training ($N_C=60$ in Omniglot and $N_C=30$ in Market1501), which leads to better performances as empirically observed in~\citet{snell2017prototypical}.
Other hyper-parameters are set to be the same as training with the triplet loss.

\begin{figure}[htbp]
    \centering
    \includegraphics[width=1.0\linewidth]{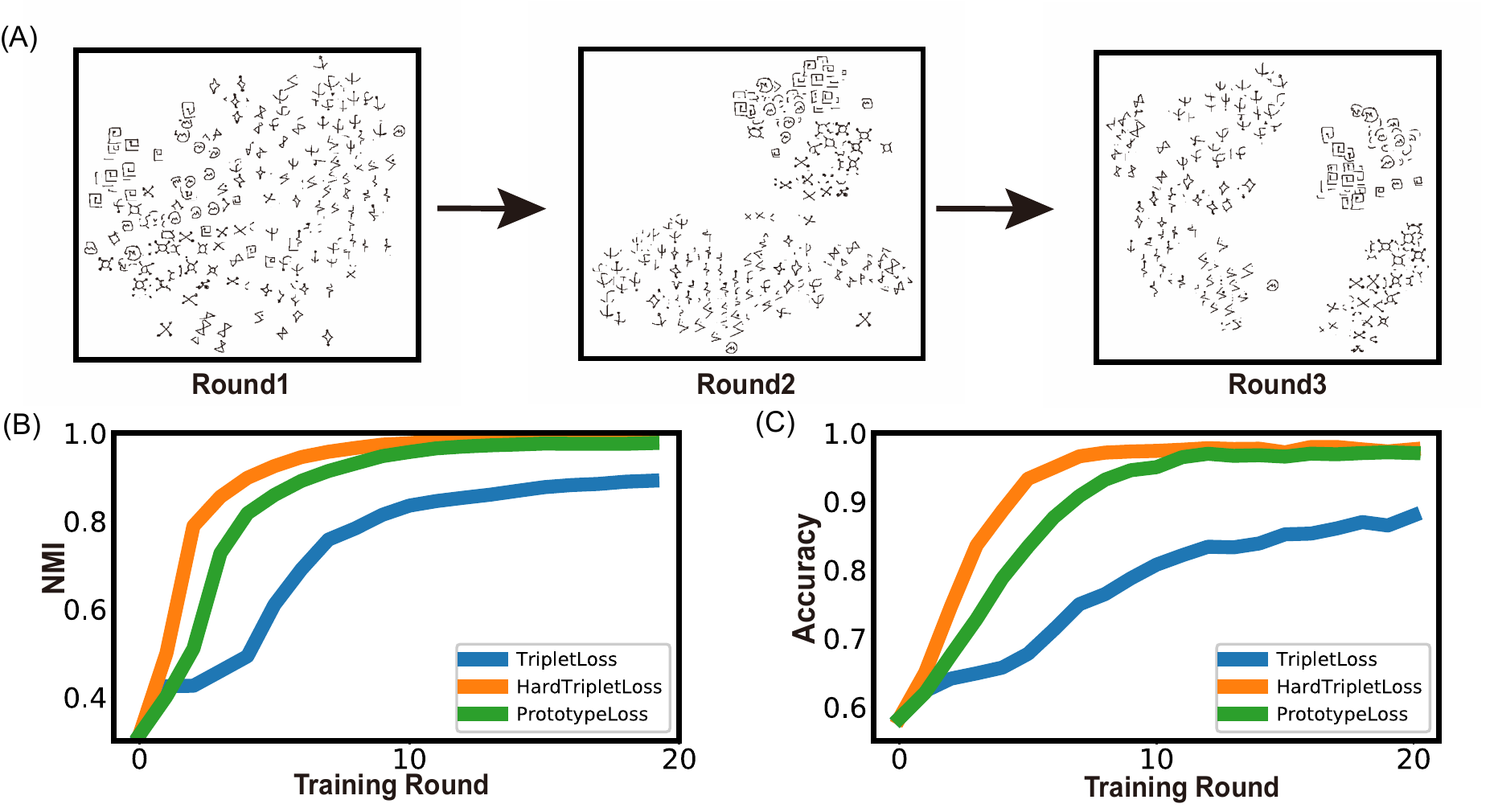}
    \caption{Behaviors of progressive clustering. 
    (A) Visualizing clustering results over training rounds by
    T-SNE. $10$ characters from the Futurama alphabets in the Omniglot dataset were selected. (B) NMI vs. training round. (C) Classification accuracy vs. training round.}
    \label{fig-nmi-accu-cluster}
\end{figure}

\subsection{Performance of Progressive Clustering}

We first check the behavior of progressive clustering via 
visualizing $10$ hand-written characters from the Futurama alphabets in the Omniglot dataset using T-SNE~\citep{maaten2008visualizing}.
Overall, we observe that as learning progresses, the organization of data points is improved continuously, indicating that our model ``discovers" the underlying structure of data gradually. 
As illustrated in Fig.~\ref{fig-nmi-accu-cluster}A,
initially, all data points are intertwined with each other and no structure exists. Over training, clusters gradually emerge, in the sense of that data points from the same class are grouped together and the margins between different classes are enlarged. 
We quantitatively measure the clustering quality by computing the Normalized Mutual Information (NMI) between real labels $\{\hat{y}_i\}^t$ (i.e., the ground truth) and pseudo labels $\{\widetilde{y}_i\}^t$,
which is given by,
\begin{equation}
     NMI\left(\{\hat{y}_i\}^t,\{\widetilde{y}_i\}^t\right)=\frac{I(\{\hat{y}_i\}^t,\{\widetilde{y}_i\}^t)}{\sqrt{H(\{\hat{y}_i\}^t)H(\{\widetilde{y}_i\}^t)}},
\end{equation}
where $I(\cdot, \cdot)$ is the mutual information between $\{\hat{y}_i\}^t$ and $\{\widetilde{y}_i\}^t$, and
$H(\cdot)$ the entropy. 
The value of NMI lies in $[0,1]$, with $1$ standing for perfect alignment between two sets. Note that NMI is independent of the permutation of labeling orders. 
As shown in Fig.~\ref{fig-nmi-accu-cluster}B, the value of NMI
increases with the training round and gradually reaches to a high value close to $1$. Remarkably, the value of NMI well predicts the classification accuracy of the learning model (comparing Fig.~\ref{fig-nmi-accu-cluster}B and 3C). 
These results strongly suggest that   
the combination of progressive clustering and episodic training in our model is able to discover the underlying structure of data manifold and extract the representative features of data points necessary for the few-shot classification task.

\subsection{Results on Omniglot}\label{omniglot-results}
Table~\ref{table-omniglot} presents the performances of our model
on the Omniglot dataset compared with other methods.  
We note that using the triplet loss, our model already outperforms other state-of-the-art unsupervised few-shot learning methods, including CACTUs~\citep{hsu2018unsupervised}, UMTRA~\citep{khodadadeh2018unsupervised}, and AAL~\citep{antoniou2019assume}, to a large extend.
Using the prototype loss, the performance of our model is further improved. The best performance of our model is achieved when 
using  the hardtriplet loss. Remarkably, the best performance of our model approaches to that of two supervised models, which are the upper bounds for unsupervised methods.

\begin{table}[htbp]
\setlength{\tabcolsep}{3.8mm}
\begin{center}
\begin{tabular}{ccccc}
  \hline
  & \multicolumn{2}{c}{\textbf{5-way Acc.}} & \multicolumn{2}{c}{\textbf{20-way Acc.}} \\
  & 1-shot & 5-shot & 1-shot & 5-shot\\
  \hline
  UMTRA~\citep{khodadadeh2018unsupervised} & 77.80 & 92.74 & 62.20 & 77.50 \\ 
  CACTUs-MAML~\citep{hsu2018unsupervised} & 68.84 & 87.78 & 48.09 & 73.36 \\
  CACTUs-ProtNets~\citep{hsu2018unsupervised} & 68.12 & 83.58 & 47.75 & 66.27 \\
  AAL-MAML++~\citep{antoniou2019assume} & 88.40 & 97.96 & 70.21 & 88.32 \\
  AAL-ProtoNets~\citep{antoniou2019assume} & 84.66 & 89.14 & 68.79 & 74.28 \\
  \hline
  UFLST-Tripletloss & 88.68 & 96.65 & 73.21 & 90.11 \\
  UFLST-Prototypeloss & 96.51 & \textbf{99.23} & 90.27 & 97.22 \\
  UFLST-HardTripletloss & \textbf{97.03} & 99.19 & \textbf{91.28} & \textbf{97.37} \\
  \hline
  MAML~\citep{finn2017model} (Supervised) & 98.7 & 99.9 & 95.8 & 98.9 \\
  ProtoNets~\citep{snell2017prototypical} (Supervised) & 98.8 & 99.7 & 96.0 & 98.9 \\
  \hline
\end{tabular}
\end{center}
\caption{Performances of different unsupervised few-shot learning models on Omniglot under different settings.}
\label{table-omniglot}
\end{table}

\subsection{Results on Mini-ImageNet}

\begin{table}[h]
\setlength{\tabcolsep}{1.2mm}
\begin{center}
\begin{tabular}{ccccc}
  \hline 
  & (5,1) & (5,5) & (5,20) & (5, 50)\\
  \hline
  Training fram scratch & 25.17 & 33.90 &39.56& 41.45 \\
  \hline
  BiGAN knn-nearest neighbors &25.56 & 31.10 & 37.31 & 43.60\\
  BiGAN linear classifier& 27.08& 33.91& 44.00 &50.41\\
  BiGAN MLP with dropout &22.91 &29.06& 40.06 &48.36\\
  BiGAN cluster matching &24.63 &29.49 &33.89 &36.13\\
  BiGAN CACTUs MAML& 36.24& 51.28& 61.33& 66.91\\
  BiGAN CACTUs ProtoNets &36.62& 50.16 &59.56& 63.27\\
  \hline
  DeepCluster knn-nearest neighbors& 28.90 &42.25 &56.44& 63.90\\
  DeepCluster linear classifier& 29.44& 39.79& 56.19& 65.28\\
  DeepCluster MLP with dropout& 29.03& 39.67& 52.71& 60.95\\
  DeepCluster cluster matching &22.20& 23.50& 24.97& 26.87\\
  DeepCluster CACTUs MAML &39.90 &\bf{53.97}& \bf{63.84}& \bf{69.64}\\
  DeepCluster CACTUs ProtoNets &\bf{39.18}& 53.36& 61.54& 63.55\\
  \hline
  UMTRA without data Augmentation & 26.49 & - & - & - \\
  UMTRA+Shift+random flip & 30.16 & - & - & - \\
  UMTRA+Shift+random flip +randomly change to grayscale  & 32.80 & - & - & - \\
  UMTRA+Shift+random flip+random rotation+color distortions & 35.09 &- & - & - \\
  UMTRA+AutoAugment & \bf{39.93}& \bf50.73& 61.11& 67.15 \\ 
  \hline
  AAL-MAML+++ CHV &33.06 &40.75 & - & - \\
  AAL-MAML+++ CHVR &33.21 & 40.34 & - & - \\
  AAL-MAML+++ CHV + CUT& 33.34&  39.44 & - & - \\
  AAL-MAML+++ CHV + DROP& 30.86&  40.41 & - & - \\
  AAL-MAML+++ CHVW &33.30 &46.98& - & - \\
  AAL-MAML+++ CHVWG &\bf{34.57} & \bf{49.18} & - & - \\
  AAL-MAML+++ CHVR + CUT& 33.09&  40.11 & - & - \\
  AAL-MAML+++ CHVR + DROP &31.70 & 39.38 & - & - \\
  AAL-MAML+++ CHV + DROP + CUT &31.55&  38.76 & - & - \\
  AAL-MAML+++ CHVR + DROP + CUT& 31.44&  39.87& - & - \\
  \hline
  AAL-ProtoNets+ CHV &\bf{37.67}& \bf{40.29} & - & - \\
    AAL-ProtoNets+ CHV + CUT &36.38  &40.89 & - & - \\
    AAL-ProtoNets+ CHV + CUT + DROP &33.13&  36.64 & - & - \\
    AAL-ProtoNets+ CHVR + CUT + DROP& 31.93 & 36.45 & - & - \\
    AAL-ProtoNets+ CHVR + CUT &33.92 &39.87& - & - \\
    AAL-ProtoNets+ CHV + DROP &32.12 & 36.12 & - & - \\
    AAL-ProtoNets+ CHVR + DROP& 31.13& 36.83 & - & - \\
    AAL-ProtoNets+ CHVR& 34.28& 39.83 & - & - \\
  \hline
  UFLST without data Augmentation & 33.77 & 45.03  & 53.35& 56.72\\
  \hline
  MAML~\citep{finn2017model} (Supervised) &46.81 &62.13& 71.03 & 75.54\\
  ProtoNets~\citep{snell2017prototypical} (Supervised) & 46.56 &62.29& 70.05 &72.04 \\
  \hline
\end{tabular}
\end{center}
\caption{Performances of different unsupervised few-shot learning models on Mini-ImageNet under different settings. The accuracy with std of our model is :$33.77\%\pm 0.70\%$, $45.03\% \pm 0.73\%$, $53.35\% \pm 0.59\%$, $56.72\% \pm 0.67\%$ on 5-way 1-shot, 5-way 5-shot, 5-way 20-shot, 5-way 50-shot, respectively.}
\label{table-mini}
\end{table}

Overall, training a few shot learner on the Mini-ImageNet dataset under the unsupervised setting is quite tricky. All the three aforementioned approaches adopt domain specific knowledge and data augmentation tricks in their training. For example, UMTRA uses the statistical likelihood of picking different classes for the training data of $T_i$ in case of $K = 1$ and large number of classes, and an augmentation function $T$ fors the validation data. CACTUs relies on an unsupervised feature learning algorithm to provide a statistical likelihood of difference and sameness in the training and validation data of $T_i$. 
The choice of the right augmentation function for UMTRA and AAL, the right feature embedding approach for CACTUs, and the other hyper-parameters have a strong impact on the performance. 

The model architecture trained on the Mini-ImageNet dataset is exactly the same as on the Omniglot dataset, i.e., the 4-layer convnet described in ~Sec.\ref{imple-detail}. We only report the results by training without any data augmentation. We achieve $33.77\%$ and $44.03\%$ under the 5-way 1-shot and 5-way 5-shot scenario respectively.  Compare to the model training from scratch ($25.17\%$ under the 5-way 1-shot scenario), our model has a gain of $8.6\%$. The best 5-way 1-shot accuracy in the CACTUs model is $39.18\%$. However, comparing to the CACTUs model is unfair because they used the AlexNet or the VGG16 to first learn a very good feature embedder for downstream feature clustering process, while our model is only composed of a 4-layer convenet. Both of the best results in the UMTRA model and the ALL model are acquired by using fancy data augmentations, such as shifting, random flipping, color distortions, image-Warping and image-pixel dropout (see ~\citet{khodadadeh2018unsupervised, antoniou2019assume} for more details) while we don't use any data augmentation tricks here. It is noteworthy that our model outperforms the UMTRA trained without any data augmentation to a large extent ($33.77\%$ vs. $26.49\%$). 

Compared to the results on Omniglot and Market1501, the results on the Mini-ImageNet is not the state-of-the-art. The underline reason may come from three aspects. 
(1) For a fair comparison to other unsupervised few-shot learning models, we use the 4-layer convnet. However, the in-class variations of the Mini-ImageNet is very large, which is hard for such a small network to capture the semantic meanings of images.
(2) In unsupervised learning, it is hard to choose suitable hyper-parameters, such as the clustering frequency, DBSCAN-related parameters, and the learning rate. 
(3) The ground truth for the class number of Mini-ImageNet is small 1(64 for training, 16 for validating and 20 for testing). But, for constructing episodic tasks, we prefer to over-segment the dataset, and this over-segmentation tend to assign data belonging to the same class into different clusters, leading to a degenerate performance. Our model performs very well on Omniglot and Maket1501, which may be attributed to that both datasets have large class numbers and the number of examples in each class is small. This type of dataset is very suitable for constructing episodic tasks to learn a few-shot learner. 
In our future work, we will explore more domain specific knowledge and data  augmentation strategies to improve the accuracy on the Mini-ImageNet dataset and extend our model to more datasets.

\subsection{Results on Market1501}\label{market1501-results}
We also applied our model to a real-world application on 
person Re-ID. In reality, labeled data is extremely lacking 
for person Re-ID, and unsupervised learning becomes crucial.  
Table~\ref{table-market1501} presents the performances of our model
on the benchmark datset Market1501.
There is no reported unsupervised few-shot learning result on this dataset in the literature. ~\citet{Rahimpour2018attention} report the supervised results under the 100-way 1-shot scenario. To evaluate our model, we trained a supervised model
adapted from \citet{xiong2018towards}.
We find that the model performance using the hardtriplet loss is much better than that using the prototype loss. This is due to 
that large variations in the appearance and environment of detected pedestrians lead to that noisy samples may be chosen as the prototypes, which deteriorates learning; while the hardtriplet loss focuses on correcting highly noisy examples that violate the margin and hence alleviates the problem.  Overall, we observe that our model achieves encouraging performances 
compared to the supervised method, in particular,
in the scenario of low-way classification, which suggest that
our model is feasible in practice for person Re-ID when annotated labels are not unavailable.

\begin{table}[htbp]
\setlength{\tabcolsep}{2.2mm}
\begin{center}
\begin{tabular}{ccccccc}
  \hline
  & 5-way & 10-way & 15-way & 20-way & 50-way & 100-way \\
  \hline
  UFLST-Tripetloss & 72.8 & 63.0 & 56.2 & 53.4 & 42.5 & 35.4 \\
  UFLST-Prototypeloss & 88.3 & 81.2 & 75.8 & 73.0 & 62.5 & 54.0 \\
  UFLST-HardTripletloss & \textbf{91.4} & \textbf{86.9} & \textbf{81.6} & \textbf{80.4} & \textbf{70.1} & \textbf{62.1} \\
  \hline
  Our supervised model & 96.8 & 94.7 & 92.5 & 91.1 & 83.7 & 77.3 \\
  ARM~\citep{Rahimpour2018attention} & - & - & - & - & - & 76.99 \\
  \hline
\end{tabular}
\end{center}
\caption{Performances of our model on Market1501 with different settings. The supervised model is adapted from \citet{xiong2018towards}.
Only 1-shot learning is considered to mimic the typical single query condition in person Re-ID applications.}
\label{table-market1501}
\end{table}

\subsection{Effect of the size of the unlabeled dataset}\label{omniglot-classes}
We also evaluate how our model relies on the number of unlabeled examples. Table~\ref{table-abalation} presents the results
on the Omniglot dataset with varying number of training examples.
Overall, the model performance is improved when
the number of training examples increases.
Notably, by using only a quarter of the unlabeled data,
our model already achieves 
performances comparable to other unsupervised few-shot learning methods (comparing UFLST-300 with those in
Table~\ref{table-omniglot}). This demonstrates the feasibility of our model when the number of unlabeled examples is not large. 

\begin{table}[htbp]
    \setlength{\tabcolsep}{7.5mm}
    \centering
    \begin{tabular}{ccccc}
        \hline
        & \multicolumn{2}{c}{\textbf{5-way Acc.}} & \multicolumn{2}{c}{\textbf{20-way Acc.}} \\
        Number of Classes & 1-shot & 5-shot & 1-shot & 5-shot\\
        \hline
          UFLST-200 & 82.83 & 92.97 & 65.85 & 83.73 \\
          UFLST-300 & 86.03 & 95.05 & 70.52 & 87.60 \\
          UFLST-400 & 91.30 & 97.27 & 78.64 & 92.50 \\
          UFLST-500 & 95.27 & 98.86 & 87.02 & 96.05 \\
          UFLST-1200 & 97.03 & 99.19 & 91.28 & 97.37 \\
        \hline
    \end{tabular}
    \caption{Performances of our model on Omniglot using different numbers of unlabeled training examples. The hardtriplet loss is used.}
    \label{table-abalation}
\end{table}

\section{Discussion}
In this study, we have proposed an unsupervised model UFLST for few-shot learning via self-training. 
The model consists of two processes, progressive clustering and episodic training, which are executed iteratively. 
Other unsupervised methods also consider the two processes, but they
are performed separately, in term of that unsupervised clustering
for feature extraction is accomplished before applying episodic learning. This separation has a shortcoming, since there is no guarantee that the extract features by unsupervised clustering are suitable for the followed few-shot learning. Here, our model carries out the two processes in an alternate manner, which allows them to facilitate with each other, such that feature representation and model generalization are optimized concurrently, and 
eventually it produces a high quality few-shot learner. To our knowledge, our work is the first one that integrates progressive clustering and episodic training for unsupervised few-shot learning. 

On the Omniglot dataset, our model outperforms other state-of-the-art unsupervised few-shot learning methods to large extend and approaches to the performances of supervised modes. 
On the MIni-ImageNey dataset, our model achieves comparable results with previous unsupervised few-shot learning models.
On the Market1501 dataset, our model also achieves encouraging performances compared to a supervised method. The high effectiveness of our model makes us think about why it works. Few-shot learning in essence is to extract good representations of data suitable for prediction by 
using very few training examples. To resolve this challenge, the episodic learning paradigm aims to create a set of episodic few-shot learning scenarios having the same setting as the testing environment, so that the model learns to extract good feature representations that are transferable to novel but related tasks. To this end, the real labels of data are helpful but not essential, and we can construct pseudo-labeled examples to train the model. But crucially,
as demonstrated by this study, the construction of pseudo-labeled examples must go along with the episodic training, so that the extracted features of data really matches the few-shot learning task. Notably, this unsupervised and progressive way of learning agrees with the nature of human on few-shot learning. 

\bibliography{main}

\begin{thebibliography}{33}
\providecommand{\natexlab}[1]{#1}
\providecommand{\url}[1]{\texttt{#1}}
\expandafter\ifx\csname urlstyle\endcsname\relax
  \providecommand{\doi}[1]{doi: #1}\else
  \providecommand{\doi}{doi: \begingroup \urlstyle{rm}\Url}\fi

\bibitem[Andrychowicz et~al.(2016)Andrychowicz, Denil, Gomez, Hoffman, Pfau,
  Schaul, Shillingford, and De~Freitas]{andrychowicz2016learning}
Marcin Andrychowicz, Misha Denil, Sergio Gomez, Matthew~W Hoffman, David Pfau,
  Tom Schaul, Brendan Shillingford, and Nando De~Freitas.
\newblock Learning to learn by gradient descent by gradient descent.
\newblock In \emph{Advances in neural information processing systems}, pp.\
  3981--3989, 2016.

\bibitem[Antoniou \& Storkey(2019)Antoniou and Storkey]{antoniou2019assume}
Antreas Antoniou and Amos Storkey.
\newblock Assume, augment and learn: Unsupervised few-shot meta-learning via
  random labels and data augmentation.
\newblock \emph{arXiv preprint arXiv:1902.09884}, 2019.

\bibitem[Caron et~al.(2018)Caron, Bojanowski, Joulin, and Douze]{caron2018deep}
Mathilde Caron, Piotr Bojanowski, Armand Joulin, and Matthijs Douze.
\newblock Deep clustering for unsupervised learning of visual features.
\newblock In \emph{Proceedings of the European Conference on Computer Vision
  (ECCV)}, pp.\  132--149, 2018.

\bibitem[de~Sa(1994)]{de1994learning}
Virginia~R de~Sa.
\newblock Learning classification with unlabeled data.
\newblock In \emph{Advances in neural information processing systems}, pp.\
  112--119, 1994.

\bibitem[Dupoux(2018)]{dupoux2018cognitive}
Emmanuel Dupoux.
\newblock Cognitive science in the era of artificial intelligence: A roadmap
  for reverse-engineering the infant language-learner.
\newblock \emph{Cognition}, 173:\penalty0 43--59, 2018.

\bibitem[Ester et~al.(1996)Ester, Kriegel, Sander, Xu,
  et~al.]{ester1996density}
Martin Ester, Hans-Peter Kriegel, J{\"o}rg Sander, Xiaowei Xu, et~al.
\newblock A density-based algorithm for discovering clusters in large spatial
  databases with noise.
\newblock In \emph{Kdd}, volume~96, pp.\  226--231, 1996.

\bibitem[Fan et~al.(2018)Fan, Zheng, Yan, and Yang]{fan2018unsupervised}
Hehe Fan, Liang Zheng, Chenggang Yan, and Yi~Yang.
\newblock Unsupervised person re-identification: Clustering and fine-tuning.
\newblock \emph{ACM Transactions on Multimedia Computing, Communications, and
  Applications (TOMM)}, 14\penalty0 (4):\penalty0 83, 2018.

\bibitem[Finn et~al.(2017)Finn, Abbeel, and Levine]{finn2017model}
Chelsea Finn, Pieter Abbeel, and Sergey Levine.
\newblock Model-agnostic meta-learning for fast adaptation of deep networks.
\newblock In \emph{Proceedings of the 34th International Conference on Machine
  Learning-Volume 70}, pp.\  1126--1135. JMLR. org, 2017.

\bibitem[Hermans et~al.(2017)Hermans, Beyer, and Leibe]{hermans2017defense}
Alexander Hermans, Lucas Beyer, and Bastian Leibe.
\newblock In defense of the triplet loss for person re-identification.
\newblock \emph{arXiv preprint arXiv:1703.07737}, 2017.

\bibitem[Hsu et~al.(2018)Hsu, Levine, and Finn]{hsu2018unsupervised}
Kyle Hsu, Sergey Levine, and Chelsea Finn.
\newblock Unsupervised learning via meta-learning.
\newblock \emph{arXiv preprint arXiv:1810.02334}, 2018.

\bibitem[Khodadadeh et~al.(2018)Khodadadeh, B{\"o}l{\"o}ni, and
  Shah]{khodadadeh2018unsupervised}
Siavash Khodadadeh, Ladislau B{\"o}l{\"o}ni, and Mubarak Shah.
\newblock Unsupervised meta-learning for few-shot image and video
  classification.
\newblock \emph{arXiv preprint arXiv:1811.11819}, 2018.

\bibitem[Lake et~al.(2015)Lake, Salakhutdinov, and Tenenbaum]{lake2015human}
Brenden~M Lake, Ruslan Salakhutdinov, and Joshua~B Tenenbaum.
\newblock Human-level concept learning through probabilistic program induction.
\newblock \emph{Science}, 350\penalty0 (6266):\penalty0 1332--1338, 2015.

\bibitem[Liu et~al.(2018)Liu, Lee, Park, Kim, Yang, Hwang, and
  Yang]{liu2018learning}
Yanbin Liu, Juho Lee, Minseop Park, Saehoon Kim, Eunho Yang, Sung~Ju Hwang, and
  Yi~Yang.
\newblock Learning to propagate labels: Transductive propagation network for
  few-shot learning.
\newblock \emph{arXiv preprint arXiv:1805.10002}, 2018.

\bibitem[Maaten \& Hinton(2008)Maaten and Hinton]{maaten2008visualizing}
Laurens van~der Maaten and Geoffrey Hinton.
\newblock Visualizing data using t-sne.
\newblock \emph{Journal of machine learning research}, 9\penalty0
  (Nov):\penalty0 2579--2605, 2008.

\bibitem[McClosky et~al.(2006)McClosky, Charniak, and
  Johnson]{mcclosky2006effective}
David McClosky, Eugene Charniak, and Mark Johnson.
\newblock Effective self-training for parsing.
\newblock In \emph{Proceedings of the main conference on human language
  technology conference of the North American Chapter of the Association of
  Computational Linguistics}, pp.\  152--159. Association for Computational
  Linguistics, 2006.

\bibitem[Mishra et~al.(2017)Mishra, Rohaninejad, Chen, and
  Abbeel]{mishra2017simple}
Nikhil Mishra, Mostafa Rohaninejad, Xi~Chen, and Pieter Abbeel.
\newblock A simple neural attentive meta-learner.
\newblock \emph{arXiv preprint arXiv:1707.03141}, 2017.

\bibitem[Nichol \& Schulman(2018)Nichol and Schulman]{nichol2018reptile}
Alex Nichol and John Schulman.
\newblock Reptile: a scalable metalearning algorithm.
\newblock \emph{arXiv preprint arXiv:1803.02999}, 2, 2018.

\bibitem[Qin et~al.(2011)Qin, Gammeter, Bossard, Quack, and
  Van~Gool]{qin2011hello}
Danfeng Qin, Stephan Gammeter, Lukas Bossard, Till Quack, and Luc Van~Gool.
\newblock Hello neighbor: Accurate object retrieval with k-reciprocal nearest
  neighbors.
\newblock In \emph{CVPR 2011}, pp.\  777--784. IEEE, 2011.

\bibitem[Rahimpour \& Qi(2018)Rahimpour and Qi]{Rahimpour2018attention}
Alireza Rahimpour and Hairong Qi.
\newblock Attention-based few-shot person re-identification using meta
  learning.
\newblock \emph{CoRR}, abs/1806.09613, 2018.
\newblock URL \url{http://arxiv.org/abs/1806.09613}.

\bibitem[Ravi \& Larochelle(2016)Ravi and Larochelle]{ravi2016optimization}
Sachin Ravi and Hugo Larochelle.
\newblock Optimization as a model for few-shot learning.
\newblock 2016.

\bibitem[Ren et~al.(2018)Ren, Triantafillou, Ravi, Snell, Swersky, Tenenbaum,
  Larochelle, and Zemel]{ren2018meta}
Mengye Ren, Eleni Triantafillou, Sachin Ravi, Jake Snell, Kevin Swersky,
  Joshua~B Tenenbaum, Hugo Larochelle, and Richard~S Zemel.
\newblock Meta-learning for semi-supervised few-shot classification.
\newblock \emph{arXiv preprint arXiv:1803.00676}, 2018.

\bibitem[Rusu et~al.(2018)Rusu, Rao, Sygnowski, Vinyals, Pascanu, Osindero, and
  Hadsell]{rusu2018meta}
Andrei~A Rusu, Dushyant Rao, Jakub Sygnowski, Oriol Vinyals, Razvan Pascanu,
  Simon Osindero, and Raia Hadsell.
\newblock Meta-learning with latent embedding optimization.
\newblock \emph{arXiv preprint arXiv:1807.05960}, 2018.

\bibitem[Snell et~al.(2017)Snell, Swersky, and Zemel]{snell2017prototypical}
Jake Snell, Kevin Swersky, and Richard Zemel.
\newblock Prototypical networks for few-shot learning.
\newblock In \emph{Advances in Neural Information Processing Systems}, pp.\
  4077--4087, 2017.

\bibitem[Song et~al.(2018)Song, Wang, Zhang, Du, Zhang, Huang, and
  Wang]{song2018unsupervised}
Liangchen Song, Cheng Wang, Lefei Zhang, Bo~Du, Qian Zhang, Chang Huang, and
  Xinggang Wang.
\newblock Unsupervised domain adaptive re-identification: Theory and practice.
\newblock \emph{arXiv preprint arXiv:1807.11334}, 2018.

\bibitem[Sung et~al.(2018)Sung, Yang, Zhang, Xiang, Torr, and
  Hospedales]{sung2018learning}
Flood Sung, Yongxin Yang, Li~Zhang, Tao Xiang, Philip~HS Torr, and Timothy~M
  Hospedales.
\newblock Learning to compare: Relation network for few-shot learning.
\newblock In \emph{Proceedings of the IEEE Conference on Computer Vision and
  Pattern Recognition}, pp.\  1199--1208, 2018.

\bibitem[Vinyals et~al.(2016)Vinyals, Blundell, Lillicrap, Wierstra,
  et~al.]{vinyals2016matching}
Oriol Vinyals, Charles Blundell, Timothy Lillicrap, Daan Wierstra, et~al.
\newblock Matching networks for one shot learning.
\newblock In \emph{Advances in neural information processing systems}, pp.\
  3630--3638, 2016.

\bibitem[Wang et~al.(2014)Wang, Zhang, Klein, Levi, and Yu]{wang2014vernier}
Rui Wang, Jun-Yun Zhang, Stanley~A Klein, Dennis~M Levi, and Cong Yu.
\newblock Vernier perceptual learning transfers to completely untrained retinal
  locations after double training: A “piggybacking” effect.
\newblock \emph{Journal of Vision}, 14\penalty0 (13):\penalty0 12--12, 2014.

\bibitem[Wang et~al.(2018)Wang, Girshick, Hebert, and Hariharan]{wang2018low}
Yu-Xiong Wang, Ross Girshick, Martial Hebert, and Bharath Hariharan.
\newblock Low-shot learning from imaginary data.
\newblock In \emph{Proceedings of the IEEE Conference on Computer Vision and
  Pattern Recognition}, pp.\  7278--7286, 2018.

\bibitem[Weinberger \& Saul(2009)Weinberger and Saul]{weinberger2009distance}
Kilian~Q Weinberger and Lawrence~K Saul.
\newblock Distance metric learning for large margin nearest neighbor
  classification.
\newblock \emph{Journal of Machine Learning Research}, 10\penalty0
  (Feb):\penalty0 207--244, 2009.

\bibitem[Xie et~al.(2016)Xie, Girshick, and Farhadi]{xie2016unsupervised}
Junyuan Xie, Ross Girshick, and Ali Farhadi.
\newblock Unsupervised deep embedding for clustering analysis.
\newblock In \emph{International conference on machine learning}, pp.\
  478--487, 2016.

\bibitem[Xiong et~al.(2018)Xiong, Xiao, Cao, Gong, Fang, and
  Zhou]{xiong2018towards}
Fu~Xiong, Yang Xiao, Zhiguo Cao, Kaicheng Gong, Zhiwen Fang, and Joey~Tianyi
  Zhou.
\newblock Towards good practices on building effective cnn baseline model for
  person re-identification.
\newblock \emph{arXiv preprint arXiv:1807.11042}, 2018.

\bibitem[Zheng et~al.(2015)Zheng, Shen, Tian, Wang, Wang, and
  Tian]{zheng2015scalable}
Liang Zheng, Liyue Shen, Lu~Tian, Shengjin Wang, Jingdong Wang, and Qi~Tian.
\newblock Scalable person re-identification: A benchmark.
\newblock In \emph{Proceedings of the IEEE international conference on computer
  vision}, pp.\  1116--1124, 2015.

\bibitem[Zhong et~al.(2017)Zhong, Zheng, Cao, and Li]{zhong2017re}
Zhun Zhong, Liang Zheng, Donglin Cao, and Shaozi Li.
\newblock Re-ranking person re-identification with k-reciprocal encoding.
\newblock In \emph{Proceedings of the IEEE Conference on Computer Vision and
  Pattern Recognition}, pp.\  1318--1327, 2017.

\end{thebibliography}
\bibliographystyle{iclr2020_conference}

\end{document}